\documentclass[letterpaper, 10 pt, conference]{ieeeconf}  
\usepackage{enumerate}

\usepackage{etoolbox}
\usepackage{covington}
\usepackage{graphicx}
\usepackage{latexsym}
\usepackage{amssymb}
\usepackage{amsmath}
\usepackage{caption}
\usepackage{subcaption}
\usepackage{mwe}
\usepackage{thmtools}
\usepackage{forloop}
\usepackage[ruled,vlined,linesnumbered]{algorithm2e}
\usepackage{paralist}
\usepackage{multirow}
\usepackage{array}
\usepackage[dvipsnames]{xcolor}
\usepackage{nicefrac}
\usepackage{pgfplots}
\usepackage{siunitx}
\usepackage{pgfplotstable}
\usepackage{booktabs}
\usepackage{numprint}
\usepackage{pifont}
\usepackage{tikz}
\usepackage{todonotes}
\usepackage[colorlinks=true, citecolor=blue, linkcolor=black]{hyperref}

\IEEEoverridecommandlockouts                              

\newtheorem{theorem}{Theorem}

\newcommand{\defcal}[1]{\expandafter\newcommand\csname c#1\endcsname{{\mathcal{#1}}}}
\newcommand{\defbb}[1]{\expandafter\newcommand\csname b#1\endcsname{{\mathbb{#1}}}}
\newcommand{\defvec}[1]{\expandafter\newcommand\csname v#1\endcsname{{\mathbf{#1}}}}
\newcommand{\defmat}[1]{\expandafter\newcommand\csname m#1\endcsname{{\mathbf{#1}}}}
\newcounter{calBbCounter}
\forLoop{1}{26}{calBbCounter}{
    \edef\capitalletter{\Alph{calBbCounter}}
		\edef\letter{\alph{calBbCounter}}
    \expandafter\defcal\capitalletter
		\expandafter\defbb\capitalletter
		\expandafter\defvec\letter
}

\newif\ifcomments
\commentstrue
\ifcomments\newcommand{\comments}[1]{#1}\else\newcommand{\comments}[1]{}\fi

\title{\LARGE \bf ORBSLAM3-Enhanced Autonomous Toy Drones: Pioneering Indoor Exploration}

\author{\href{https://scholar.google.com/citations?user=721xaz0AAAAJ\&hl=en\&oi=ao}{\color{blue}Murad Tukan}$^{1}$ and \href{https://scholar.google.com/citations?user=Wm4eHwgAAAAJ\&hl=en}{\color{blue}Fares Fares}$^{2}$ and Yotam Grufinkle$^{3}$ and Ido Talmor$^{4}$ and  \\ and Loay Mualem$^{5}$ and \href{https://scholar.google.com/citations?hl=en\&user=DTthB48AAAAJ}{\color{blue}Vladimir Braverman}$^{6}$ and \href{https://scholar.google.com/citations?user=67QZN0gAAAAJ\&hl=en}{\color{blue} Dan Feldman}$^{7}$
\thanks{$^{1}$ Murad Tukan is with DataHeroes, Israel {\tt\small murad@dataheroes.ai}}
\thanks{$^{2}$ Fares Fares is with the Robotics and Big Data lab, University of Haifa, Haifa, Israel {\tt\small faresfares.cs@gmail.com}}
\thanks{$^{3}$ Yotam Grunfinkle is with the Robotics and Big Data lab, University of Haifa, Haifa, Israel {\tt\small yotam706@gmail.com }}
\thanks{$^{4}$ Ido Talmor is with the Robotics and Big Data lab, University of Haifa, Haifa, Israel {\tt\small ifw.talmor@gmail.com}}
\thanks{$^{5}$ Loay Mualem is with the Computer Science Department, University of Haifa, Haifa, Israel, and with DataHeroes, Israel {\tt\small loaymua@gmail.com}}
\thanks{$^{6}$ Vladimir Braverman is with the Computer Science Department, Rice University, Houston Texas, USA {\tt\small vb21@rice.edu}}
\thanks{$^{7}$ Dan Feldman is with the Robotics and Big Data lab, University of Haifa, Haifa, Israel {\tt\small dannyf.post@gmail.com }}
\thanks{$\ast$ This work has been submitted to the IEEE for possible publication.
Copyright may be transferred without notice, after which this version may
no longer be accessible.}}

\newcommand{\exitfind}{\textsc{Exit-Finder}}
\newcommand{\term}[1]{\left( #1 \right)}
\newcommand{\br}[1]{\left\lbrace #1 \right\rbrace}
\newcommand{\abs}[1]{\left| #1 \right|}
\newcommand{\norm}[1]{\left\| #1 \right\|_2}

\usepackage{algorithmic}
\usepackage{eqparbox}

\newcommand{\say}[1]{``#1"}

\begin{document}

\maketitle
\thispagestyle{empty}
\pagestyle{empty}

\begin{abstract}
Navigating toy drones through uncharted GPS-denied indoor spaces poses significant difficulties due to their reliance on GPS for location determination. In such circumstances, the necessity for achieving proper navigation is a primary concern. In response to this formidable challenge, we introduce a real-time autonomous indoor exploration system tailored for drones equipped with a monocular \emph{RGB} camera. 

Our system utilizes \emph{ORB-SLAM3}, a state-of-the-art vision feature-based SLAM, to handle both the localization of toy drones and the mapping of unmapped indoor terrains. Aside from the practicability of \emph{ORB-SLAM3}, the generated maps are represented as sparse point clouds, making them prone to the presence of outlier data. To address this challenge, we propose an outlier removal algorithm with provable guarantees. Furthermore, our system incorporates a novel exit detection algorithm, ensuring continuous exploration by the toy drone throughout the unfamiliar indoor environment. We also transform the sparse point to ensure proper path planning using existing path planners.

To validate the efficacy and efficiency of our proposed system, we conducted offline and real-time experiments on the autonomous exploration of indoor spaces. The results from these endeavors demonstrate the effectiveness of our methods.
\end{abstract}

\section{BACKGROUND}


Unmanned aerial vehicles (UAVs) have gained popularity over recent years for their wide array of applications, including aerial photography~\cite{li2012design}, surveillance~\cite{semsch2009autonomous}, environmental monitoring~\cite{asadzadeh2022uav}, disaster response~\cite{hildmann2019using}, and search and rescue missions~\cite{erdos2013experimental}. 

An essential goal within the realm of UAVs (Unmanned Aerial Vehicles) is to operate in close proximity to objects and navigate through environments that pose challenges for both humans and traditional aircraft. In these scenarios, compact UAVs, such as Toy-Sized~\cite{jubran2022newton} ones, play a significant role. Such agents usually weigh less than $250$ grams, due to regularization and safety issues, equipped with a monocular camera, for example, a Tello drone. Nevertheless, achieving autonomous exploration in GPS-denied unknown environments remains a formidable challenge for such agents. To accommodate proper exploration, one needs to account for (i) mapping the environment and (ii) using a proper path planner to move the agent from its current position to some goal point. 

\subsection{Mapping the Environment} Employing an autonomous exploration system for toy drones requires mapping the environment for proper localization and navigation through the environment. Next, we discuss widely used robotic mapping systems.

\textbf{\emph{LiDAR}-based systems.} \emph{LiDAR} sensors are technically laser beams measuring distances that can be utilized to produce a $3$D map (or equivalently $2$D map) of the surroundings~\cite{gao2018autonomous, camargo2019mobile, ismail2022exploration}. These systems not only generate precise maps of the surroundings but also perform simultaneous robot localization within the map. One significant drawback of such systems is that the cost of involved \emph{LiDAR} sensors can be relatively high, potentially restricting their use in certain applications.

\textbf{Vision-\emph{SLAM}-based systems.} Unlike \emph{LiDAR}-generated maps, \emph{SLAM}-generated maps can be represented in various forms, such as $2$D or $3$D grids, point clouds, or feature-based representations. Such maps are being created and updated in an incremental fashion as the agent moves through them. These \emph{SLAM}-systems have shown promising results in the field of indoor exploration. For example, \emph{RGB-D} SLAM~\cite{henry2010rgb} was utilized in an autonomous exploration system of unknown indoor environments via a novel exploration strategy exploiting the number of features and their distribution uniformity score in $3$D~\cite{eldemiry2022autonomous}. While \emph{RGB-D} SLAM-based systems allow for more detailed and accurate mapping due to the additional depth of information, their need comes with the existence of lighting conditions or where depth information is critical (e.g., robotics for object avoidance) via the use of depth cameras. When the usage of lightweight and efficient SLAM systems for real-time applications is a requirement, such systems are not ideal. 

As for monocular RGB-camera \emph{SLAM}-based systems, \emph{LSD-SLAM}~\cite{engel2014lsd} is a natural candidate for its scalable mapping, i.e., its ability to handle environments with varying scales. This system was utilized in~\cite{von2017monocular} to estimate the trajectory of the agent as well as ensure a semi-dense representation of the environment of the agent. While such a \emph{SLAM} system can operate with minimal hardware such as a simple cheap monocular camera, the computational requirements of such a system may vary depending on the scale of the environment.

\textbf{\emph{ORB-SLAM3}}~\cite{campos2021orb}. This SLAM variant exhibits distinct advantages when compared to \emph{LSD-SLAM} and \emph{RGB-D} SLAM, particularly in terms of its robustness. It excels even in challenging scenarios marked by fluctuating lighting conditions and the presence of dynamic objects, among other complexities. Furthermore, \emph{ORB-SLAM3} stands out as a cost-effective solution since it does not necessitate the use of additional high-cost sensors like \emph{LiDAR}. Lastly, it is well-suited for applications where minimizing latency is paramount, providing a significant edge over other systems, such as \emph{LSD-SLAM}, known for its computational intensity.

\subsection{Path Planning}
Path planning represents a pivotal challenge in the field of robotics. Its primary objective is to discover a viable route from an initial state to a desired final state while ensuring that the path avoids any potential obstacles. The principal difficulty arises from the typically vast and intricate search space that needs to be explored, especially when dealing with continuous problems. A common strategy employed to address this challenge is the strategic sampling of state configurations, with the attempt to uncover a feasible path. In practice, sampling-based path planning methods, such as Rapidly-exploring Random Trees (RRTs)~\cite{lavalle1998rapidly} and Probabilistic Roadmaps (PRMs), have gained widespread popularity as effective choices for navigating this intricate terrain leading to a vast array of variants of such path planners~\cite{lavalle1998rapidly,karaman2010incremental,islam2012rrt,c2, salzman2016asymptotically,petit2021rrt,tukan2022obstacle}. Traditionally in the field of path planning, the state space is composed mainly of two sets -- the free space and the space of obstacle states where the latter is composed of fully bounded bodies (e.g., polytopes, convex bodies, etc.). In the context of \emph{ORB-SLAM3}, obstacles are not represented by fully bounded bodies, but rather a sparse set of points. Thus, this prompts the question of whether \textbf{is it possible to maneuver the obstacle through such representation while ensuring collision avoidance?}

\begin{figure}[!htb]
    \centering
    \includegraphics[width=\linewidth,height=4cm]{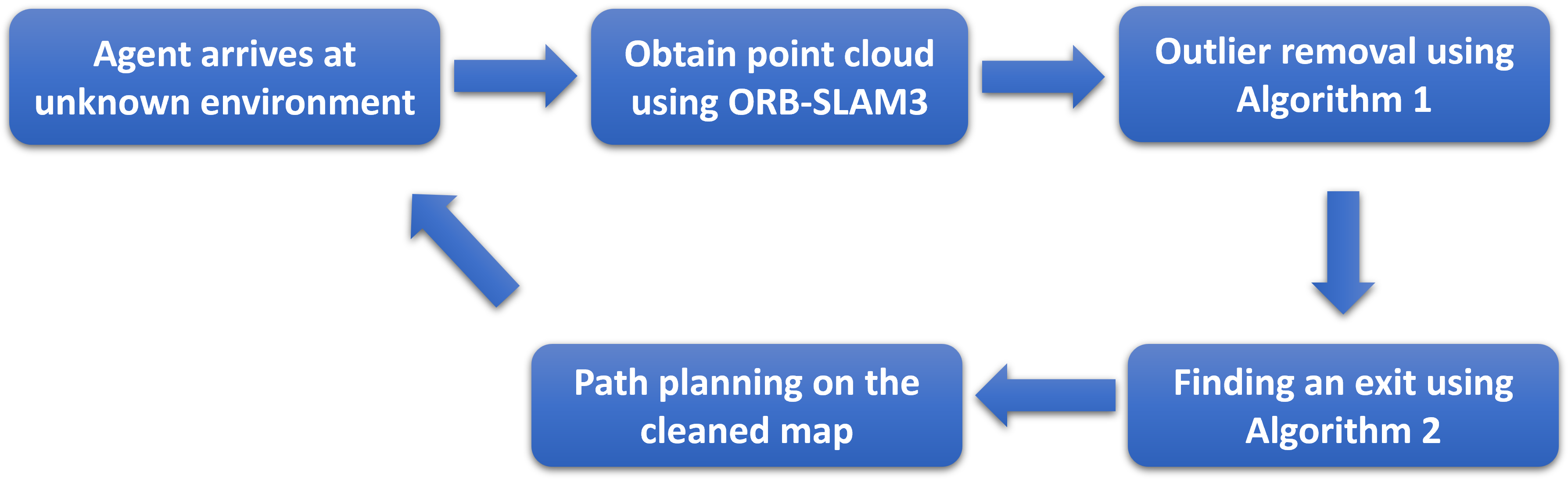}
    \caption{An overview of our system's pipeline.}
    \label{fig:erd}
\end{figure}

\section{Our Contribution}

In this work, we provide a real-time autonomous exploration system in an indoor GPS-denied environment using a toy drone equipped with a single RGB monocular camera based on \emph{ORB-SLAM3}. Problems arising towards such a goal consist of the lack of accurate positioning of the agent, the occurrence of noisy features grasped by \emph{ORB-SLAM3}~\cite{cao2020study}, and adjusting path planners to sparse point clouds. In this context, there is no clear distinction between an obstacle point and a noisy point. This raises the question:

\textbf{Is it possible to provide a real-time \emph{ORB-SLAM3} based exploration system in indoor GPS-denied environments while adapting path planners to operate with sparse representation of such maps?}

To that end, we answer the above question in the affirmative, where our contribution is four-fold:
\begin{enumerate}[(i)]
 

    \item Detecting and removing outliers (noise) via optimizing a submodular function with theoretical guarantees. 
    
    \item Developing a $2$D representation of the point cloud generated by \emph{ORB-SLAM3} for finding goal (exit) points leading to exploration of unknown environments.


    \item Transforming sparse point clouds to a composition of free and obstacle states, thus allowing path planners to handle sparse data. 

    \item Open code of our proposed system~\cite{opencode}.
\end{enumerate}




\section{SYSTEM}
\subsection{Key Idea}
Given a lightweight toy drone ($<100$gr) equipped with only a monocular RGB camera, the purpose of this work is to provide an autonomous system for state-space exploration of a low-cost drone utilizing a feature-based tracking and positioning system, namely, \emph{ORB-SLAM3}.

\textbf{\emph{ORB-SLAM3} related problems.} First notice that \emph{ORB-SLAM3} generates a sparse point cloud representing the local environment of the agent, referred to as $\mathcal{M}$. Traditionally, $\mathcal{M}$ is represented via a composition of two sets of states, one reserved for the free states (obstacle-free locations), and the other set is for obstacles that are usually defined by dense bodies. An example of such representation is an occupancy grid~\cite{elfes1989using,collins2007occupancy}. While such a representation eases the task of navigating an agent through $\mathcal{M}$, recall that in our context $\mathcal{M}$ is represented by a sparse set of points. As a result, the agent is faced with the added challenge of maneuvering through $\mathcal{M}$ for multiple reasons.

For example, sparse representations of the data make the task of distinguishing between different obstacles harder, since the barrier between pairs of close obstacles is either under-sampled or poorly represented. As a byproduct of this, a completely dense obstacle is depicted as having a sufficiently large gap or opening, resulting from the incomplete capture of essential features, allowing the agent to pass through.

Bear in mind that \emph{ORB-SLAM3} is prone to noisy samples, making the goal of navigation even harder to achieve. Such noisy samples can occur due to illumination, reflective surfaces, etc. This is in addition to the \emph{ORB-SLAM3} incapability of grasping enough features to represent either a closed window or a white blank wall. In summary, our \emph{ORB-SLAM3}-based navigation system, faces three major problems:
\begin{itemize}
    \item Existence of outliers.
    \item Finding goal points to navigate to. 
    \item Path planning through sparse point clouds (\emph{ORB-SLAM3} generated features).
\end{itemize}


\textbf{Assumptions.} To ensure autonomous navigation with \emph{ORB-SLAM3}, the only assumption we require is that any environment has enough features to be grasped by \emph{ORB-SLAM3}, in order to adequately represent the environments that our agent (e.g., drone) is passing through.

\textbf{Notations.} The following notations will be used throughout the paper. For any set of points $X \subseteq \mathbb{R}^d$, let $\abs{X}$ denote the cardinality of $X$. For any positive integer $n$, let $[n]$ denote the set $\br{1,2,3,\cdots, n}$. Finally, we let $(A,v)$ denote an affine subspace (a linear subspace that does not necessarily pass through the origin of $\mathbb{R}^3$) where $A \in \mathbb{R}^{3 \times 2}$ is an orthonormal matrix and $v \in \mathbb{R}^3$ is its translation from the origin.

The steps of our system are illustrated in Figure~\ref{fig:erd}. In the following subsections, we dive into each of the components that compose our system.

\subsection{Outlier Removal}



To address the challenge of capturing noisy features, we resort to subset selection methodologies. Note that subset selection techniques have gained much success throughout various fields, e.g., classification~\cite{munteanu2018coresets,tukan2020coresets, tukan2021svm,maalouf2023autocoreset}, clustering~\cite{bachem2018scalable,zhang2020fast,jubran2020sets,tukan2022new,tukan2023efficient}, function fitting~\cite{maalouf2022coresets}. 

Different from prior work utilizing subset selection techniques, in this paper, we utilize such methods for identifying and eliminating outliers, specifically via employing \emph{minimax optimization} for a \emph{jointly-submodular} function. 
Recently, submodular functions have gained increasing interest in many machine learning and robotics problems, e.g.,~\cite{zhou2022distributed,mualem2022using,mualem2023resolving,mualem2023submodular,shi2021communication}. In this paper, we use minimax \emph{jointly-submodular} function optimization to remove outliers in the point cloud generated by \emph{ORB-SLAM3}.



\paragraph{Minimax Submodular Optimization}
We start with the property of \emph{jointly-submodular} functions. Given an element $u$ and two ground sets $\cN_1$ and $\cN_2$, we denote by $f(u\mid S)\triangleq f(S \cup \{u\}) - f(S)$ the marginal contribution of the element $u$ to the set $S$. The function $f$ is \emph{jointly-submodular} if for every two sets $S \subseteq T \subseteq \cN_1 \cup \cN_2$ and element $u \in (\cN_1 \cup \cN_2) \setminus T$, it holds that
\[
	f(u \mid S) \geq f(u \mid T).
\]

Formally speaking, given two disjoint sets $\cN_1,\cN_2$, where each set is a copy of the point cloud generated by \emph{ORB-SLAM3}, and two sets $X \subseteq \cN_1, Y \subseteq \cN_2$, we define \emph{\say{outliers score}} as follows.
\begin{align}\label{func:img}
	f(X \cup Y)
	={} \mspace{-9mu}
\sum_{v\in \mathcal{N}\setminus X} \mspace{-4mu} \max_{u\in Y}s_{u,v}-\frac{1}{|\mathcal{N}_2|}\sum_{u\in Y}\sum_{v\in Y} s_{u,v} + \lambda\cdot|X|,
\end{align}
where $s_{u,v}$ is a \emph{\say{distance score}} defined by $s_{u,v}=1-1/d(u,v)$--here, $d(u,v)$ represents the Euclidean distance between points $u,v$. A similar distance score was used by Mualem et al.~\cite{mualem2023submodular}, who also showed that~\eqref{func:img} is a non-negative \emph{jointly-submodular} function.


The function~\eqref{func:img} consists of three terms. The first term represents a variation of the well-known Facility Location\footnote{A Facility Location function represents the cost of a particular placement of facilities as a function of transportation costs and other factors.} function, which captures the representativeness of the set $Y$. The second term ensures the diversity of the chosen points, and the last term controls the number of detected outliers. Roughly speaking, we aim to find a set of points $\cN_1\setminus X$ that is good with respect to every representative set $Y$ of some given size $k$ in the sense that every point in $Y$ is close to at least one point of $\cN_1 \setminus X$. Thus, we have to (approximately) solve the following optimization problem.
\begin{equation}
\label{eq:optimization_prob}
\min_{X\subseteq\mathcal{N}_1}\max_{\substack{Y \subseteq \cN_2\\|Y| \leq k}}f(X \cup Y)
	\enspace.    
\end{equation}

\paragraph{Method} Maximizing submodular functions is NP-hard~\cite{feige1998threshold}. However, it is possible to approximately optimize Equation~\eqref{eq:optimization_prob} using Algorithm~\ref{alg:square_root_approximation} (due to~\cite{mualem2023submodular}). See Theorem~\ref{thm:min_max_sqrt} for the (informal) theoretical guarantee of this algorithm.


\textbf{Overview of Algorithm~\ref{alg:square_root_approximation}.} First, we note that $X_0$ in our context is an empty set since $f(\emptyset) = 0$; see Line~\ref{line:sub1}. The algorithm then finds a representative set of points of size at most $k$ from the whole data by optimizing the maximization component (akin to finding a geographical $k$-center from the data), aiming to expand its exploration of inliers; see Line~\ref{line:sub3}. Specifically speaking, for every $i \in \left[\abs{\cN_1} + 1\right]$ at Lines~\ref{line:sub3}--~\ref{line:sub4}, a submodular optimization oracle is utilized to find a subset $Y_i$ containing at most $k$ items of $\cN_2$ that represent the remaining items of $\cN_1 \setminus X_{i-1}$. We then find a subset $X^\prime_i$ of items from $\cN_1$ that are close to $Y_i$, and add them to $X_{i - 1}$ to get the new set $X_i$; see Lines~\ref{line:sub5}--~\ref{line:sub10}.

\begin{algorithm}[!htb]
\DontPrintSemicolon
\caption{Iterative $X$ Growing} \label{alg:square_root_approximation}
\begin{algorithmic}[1]
\REQUIRE{A pair of sets of points $\cN_1, \cN_2 \subseteq \mathbb{R}^3$, a
\\\>submodular maximization oracle $\mathcal{O}$, and number of centers $k$.}
\ENSURE{A subset $S\subseteq\cN_1$ of inliers.}
\STATE Use an algorithm for submodular minimization to \\\>find a set $X_0 \in \arg \min_{X \subseteq \cN_1} f(X)$ \label{line:sub1}
\FOR{$i = 1$ \KwTo $\abs{\cN_1} + 1$\label{line:sub2}}
\STATE Let $\mathcal{F}_2 := \br{Y \mid Y \subseteq \cN_2, \abs{Y} \leq k}$  \label{line:sub3}
\STATE Use the oracle $\mathcal{O}$ to find $Y_i \in \cF_2$ maximizing $f(X_{i - 1} \cup Y_i)$ up to a factor of $\alpha$ among all sets \\\>in $\cF_2$\label{line:sub4}

\STATE Use an algorithm for submodular minimization to \\\>find a set $X'_i \in \arg \min_{X \subseteq \cN_1} \beta f(X \cup X_{i - 1}) + f(X \cup Y_i)$ \label{line:sub5}
\IF{$X'_i \subseteq X_{i - 1}$\label{line:sub6}}
\STATE \textbf{Return} $X_{i-1}$ \label{line:sub7}
\ELSE\label{line:sub8}
\STATE Let $X_i \gets X_{i - 1} \cup X'_i$.\label{line:sub9}
\ENDIF\label{line:sub10}
\ENDFOR\label{line:sub11}
\end{algorithmic}
\end{algorithm}

The algorithm repeats this procedure until $X^\prime_i \subseteq X_{i-1}$ (guaranteed to occur at some iteration). In our experiments, we employ the conventional Greedy Algorithm~\cite{nemhauser1978analysis,mualem2022using} as our maximization oracle $\mathcal{O}$, which ensures a guaranteed approximation of $\alpha=1-\nicefrac{1}{e}$. To visualize our Outlier Removal approach, we refer the reader to Figure~\ref{fig:clean}.

\begin{theorem}[Informal version of Theorem 3.3~\cite{mualem2023submodular}]\label{thm:min_max_sqrt}
For an appropriate choice of the parameter $\beta$, Algorithm~\ref{alg:square_root_approximation} returns a set $\hat{X} \subseteq \cN_1$ such that $\max_{Y \in \cF_2} f(\hat{X} \cup Y)$ is lower bounded by $\tau$ and upper bounded by $O(\alpha\sqrt{|\cN_1|}) \cdot \tau$, where $\tau \triangleq \min_{X \subseteq \cN_1} \max_{Y \in \cF_2} f(X \cup Y)$.
\end{theorem}

\subsection{Exit Detection}

Following the outlier removal procedure concerning the local environment $\mathcal{M}$ of the agent (as described in the previous sub-section), we aim to 
find a point of interest that leads the agent to continue exploring the unknown environments.

To achieve this goal, it is important to acknowledge that the local environment data $\mathcal{M}$ obtained through \emph{ORB-SLAM3} can be unclear in distinguishing subsets of data points representing elements like walls, windows, or entrances due to inherent inaccuracies. For example, walls are often represented by non-linear point collections in $3$D space, making it challenging to model them as linear subspaces accurately. This inaccuracy can result in significant deviations between the modeled subspaces and the actual wall structure. Furthermore, the presence of noisy data near entrances complicates the identification of exit points, whether they are open doors or elongated corridors.



\paragraph{LiDAR-based navigation system} Recall that in such systems, these challenges are typically mitigated thanks to the LiDAR method's ability to measure the time it takes for reflected light to return to the receiver, which is a key feature of LiDAR sensors. With this functionality, it is possible to construct a precise $2$D map that faithfully represents the local environment of the agent, as demonstrated in~\cite{hess2016real}, while avoiding the presence of highly inaccurate samples. In contrast, \emph{ORB-SLAM3} generates a sparse point cloud representation of the same environment, which may lack the same level of accuracy.


\paragraph{Reproducing \say{\emph{LiDAR}} maps without any additional hardware}
To alleviate this drawback, our goal is to generate a $2$D representation of the provided map that preserves a structure similar to \emph{LiDAR} maps, using solely a monocular RGB-camera and \emph{ORB-SLAM3}.
To achieve this objective, two crucial components are required: (i) the agent's position, and (ii) $\mathcal{M}$, which both are guaranteed by \emph{ORB-SLAM3}. Next, we detail our algorithm that is designed to create maps with a structure closely resembling that of \emph{LiDAR} maps. The ultimate goal is to identify a point of interest, referred to as an \say{exit point}, for the agent to navigate towards.



\begin{figure}[!htb]
    \centering
    \includegraphics[width=\linewidth, height=6cm]{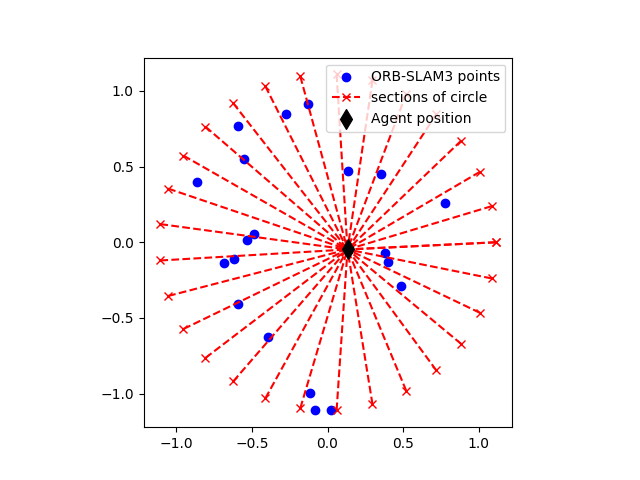}
    \caption{Example of clustering \emph{ORB-SLAM3} points to 30 groups based on their corresponding relative angle to the agent's position (highlighted by a black diamond). A point $p$ is assigned to a circle sector $i \in [30]$ if its relative angle to the agent's position is in the range $\left[ i  * 2\pi / 30, (i+1) * 2\pi /30 \right)$. Note that some of the groups might have no points at all hinging upon a possible entrance to an unexplored area.}
    \label{fig:binning}
\end{figure}

\textbf{Overview of Algorithm~\ref{alg:exit}.} Given a set of $n$ points $\mathcal{M} \subseteq \mathbb{R}^3$ representing a local map of our agent, the agent's position $x \in \mathbb{R}^3$ and an affine subspace $\term{A, v}$ which will be used to project the $3$D map to $2$D for easier processing, Algorithm~\ref{alg:exit} returns a point in $3$D that lies on the affine subspace $\term{A,v}$ representing an exit point. The algorithm starts by projecting the local map from $3$D to $2$D using the affine subspace $(A,v)$ as well as the current agent's position, namely, $x$; see Line~\ref{line:exit1}. Afterward, the angle between each point in the $2$D projected map $\mathcal{M}^\prime$ and the $2$D projected position $x^\prime$ of the agent is computed followed by creating $360$ empty groups as depicted at Lines~\ref{line:exit2}--\ref{line:exit3}. The points in $\mathcal{M}^\prime$ are then dissected to $360$ groups ($B_1,B_2,\ldots, B_{360}$) such that a point $p \in \mathcal{M}^\prime$ is assigned to a set $B_i$ for some $i \in [360]$ if the angle between $p$ and $x^\prime$ lies in the range $\left[ i \cdot \pi / 180, (i+1) \cdot \pi /180 \right)$; see Lines~\ref{line:exit5}--~\ref{line:exit13} and Figure~\ref{fig:binning}. Moving forward, a $2$D map is created such that the $\mathcal{X}$-axis represents angles between points in $\mathbb{R}^2$ and $x^\prime$. Specifically speaking, we aim to create a map of $2$D points, where a point is present at some $(x,y)$ coordinates if $x \in [0, 2\pi]$ for some $i\in [360]$ such that $x \in B_i$ and $\abs{B_i} > 1$. We then define $y$ to be the average distance between points in $B_i$ to $x^\prime$. This is depicted at Lines~\ref{line:exit14}-~\ref{line:exit20} of Algorithm~\ref{alg:exit}.

\begin{algorithm}[!htb]
    \caption{$\exitfind\term{\mathcal{M}, x, \term{A,v}}$}
    \label{alg:exit}
    \begin{algorithmic}[1]
     \REQUIRE{A set $\mathcal{M} \subseteq \mathbb{R}^3$ of $n$ points representing the local map of our agent, the position $x \in \mathbb{R}^3$ of our agent, and an affine subspace $\term{A, v}$}
    \ENSURE{An exit point $y$ that lies on the affine subspace $\term{A,v}$\\}
    \COMMENT{$2$D Projection}
    \STATE Let $\mathcal{M}^\prime := \br{A^T\term{p - v} \middle| p \in \mathcal{M}}$ and let $x^\prime := A^T(x-v)$ \label{line:exit1}
    \STATE For every $p \in \mathcal{M}^\prime$, Let $\alpha(p):= $ the radian angle between $x^\prime$ and $p$ \label{line:exit2}\\
    \COMMENT{Dissect $\mathcal{M}^\prime$ into $360$ bins}
    \STATE Let $\mathcal{B} := \br{B_i}_{i=1}^{360}$ be a set of $360$ empty bins \label{line:exit3}
    \STATE Let $r := 0$ \label{line:exit4}
    \FOR{ every point $p \in \mathcal{M}^\prime$ \label{line:exit5}}
    \STATE $r := r + \frac{1}{\abs{\mathcal{M}^\prime}} \norm{p}$ \label{line:exit6}
    \FOR{every $i \in [360]$ \label{line:exit7}}
    \IF{$\alpha\term{p} \in \left[ i \cdot \pi / 180, (i+1) \cdot \pi /180 \right)$ \label{line:exit8}}
    \STATE $B_i = B_i \cup \br{p}$ \label{line:exit9}
    \STATE break \label{line:exit10}
    \ENDIF \label{line:exit11}
    \ENDFOR \label{line:exit12}
    \ENDFOR \label{line:exit13}\\ 
    \COMMENT{$2$D map production}
    \STATE Let $L := \emptyset$ \label{line:exit14} 
    \FOR{every $i \in [360]$ \label{line:exit15}} 
    \IF{$\abs{B_i} > 1$ \label{line:exit16}}
    \STATE Let $\hat{d} := \frac{1}{\abs{B_i}}\sum_{p \in B_i} \norm{p - x^\prime}$ \label{line:exit17} 
    \STATE $L := L \cup \br{\begin{bmatrix}
        \frac{\beta}{180} \\ \hat{d}
    \end{bmatrix} \middle| \beta \in \left[ i \cdot \pi, (i+1) \cdot \pi \right)}$ \label{line:exit18}
    \ENDIF \label{line:exit19}
    \ENDFOR \label{line:exit20}\\
    \COMMENT{Finding an exit point}
    \STATE Let $s:= $ the largest segment contained in $\left[0, 2\pi \right]$ \\\>such that no point in $L$ has $\mathbf{x}$-axis coordinate in the segment \label{line:exit21}
    \STATE Let $\bar{s} := $ the median element of $s$ \label{line:exit22}
    \STATE $y^\prime := x^\prime + \begin{bmatrix}
        r\cos\term{\bar{s}}\\
        r\sin\term{\bar{s}}\\
    \end{bmatrix}$ \label{line:exit23}
    \STATE $y := Ay^\prime + v$ \label{line:exit24}
    \STATE \textbf{Return} y \label{line:exit25}

    \end{algorithmic}
\end{algorithm}

We are now able to find the angle that can lead to the exit by finding the largest continuous set of angles from the range $\left[ 0, 2\pi \right]$ such that there exists no point in $L$ that has a $\mathbf{x}$-axis coordinate in such segment. The angle of the exit point $\bar{s}$ is set to be the median of the segment $s$; see Line~\ref{line:exit22}. The exit point in $2$D is then set to be along the angle $\bar{s}$ (Line~\ref{line:exit23}) at a distance $r$ (the mean norm of the projected points $\mathcal{M}^\prime$ as presented at Line~\ref{line:exit6}). Finally, to map back to a $3$D representation, we simply project back onto the affine space in its $3$D form; see Lines~\ref{line:exit23}--\ref{line:exit24}.

\subsection{Path Planning}
\label{sec:refinement}

After obtaining an exit point on some $2$D plane in $\mathbb{R}^3$, we proceed to find a collision-free path between the agent position and the proposed exit point.

\textbf{Using the $2$D map from Algorithm~\ref{alg:exit}.} 
Recall that \emph{ORB-SLAM3}-generated maps differ from the data format typically processed by path planners. While the $2$D map we created appears suitable for path planning due to its rigid lines, it simplifies by focusing on average distances within angle ranges. This neglects individual distances between the agent and points within each angle range, potentially leading to collisions by not accounting for obstacle positions along the agent's path to the exit point. 


\textbf{From sparse to dense.} 
Unlike traditional path planners that deal with dense obstacle representations, our approach addresses sparse point-based obstacle data from \emph{ORB-SLAM3}, where clear obstacle boundaries are lacking. Using a standard path planner on such maps can produce paths that pass through obstacles due to gaps in \emph{ORB-SLAM3}'s obstacle representation. To resolve this, we propose a solution that combines $K$-means clustering and convex hull computation. After the outlier removal stage in our system, we cluster the map with a large number $K$ of clusters (e.g., $K=300$) using $K$-means, and then calculate a convex hull for each resulting cluster. This process creates our obstacle space, consisting of polytopes representing the obstacles.


With the set of obstacles above (set of convex hulls), we proceed to use the RRT path planner~\cite{lavalle1998rapidly} due to its simplicity and practicability.

\textbf{Path refinement.} The inherent randomness of RRT can lead to paths within the RRT tree that are occasionally unnecessarily complex, potentially resulting in longer routes when shorter ones are available. To tackle this problem, we devised a solution. For each sub-path of the path generated by RRT (between the agent's position and the exit point), starting at point $A$ and ending at point $B$, we assess a single edge connecting $A$ and $B$ to determine its collision status. If the edge is collision-free, we replace the sub-path between $A$ and $B$ with this single edge.

\section{EXPERIMENTAL RESULTS}
\begin{figure*}[!htb]
\begin{subfigure}[t]{0.33\textwidth}
 \centering
    \includegraphics[width=\textwidth,height=3cm]{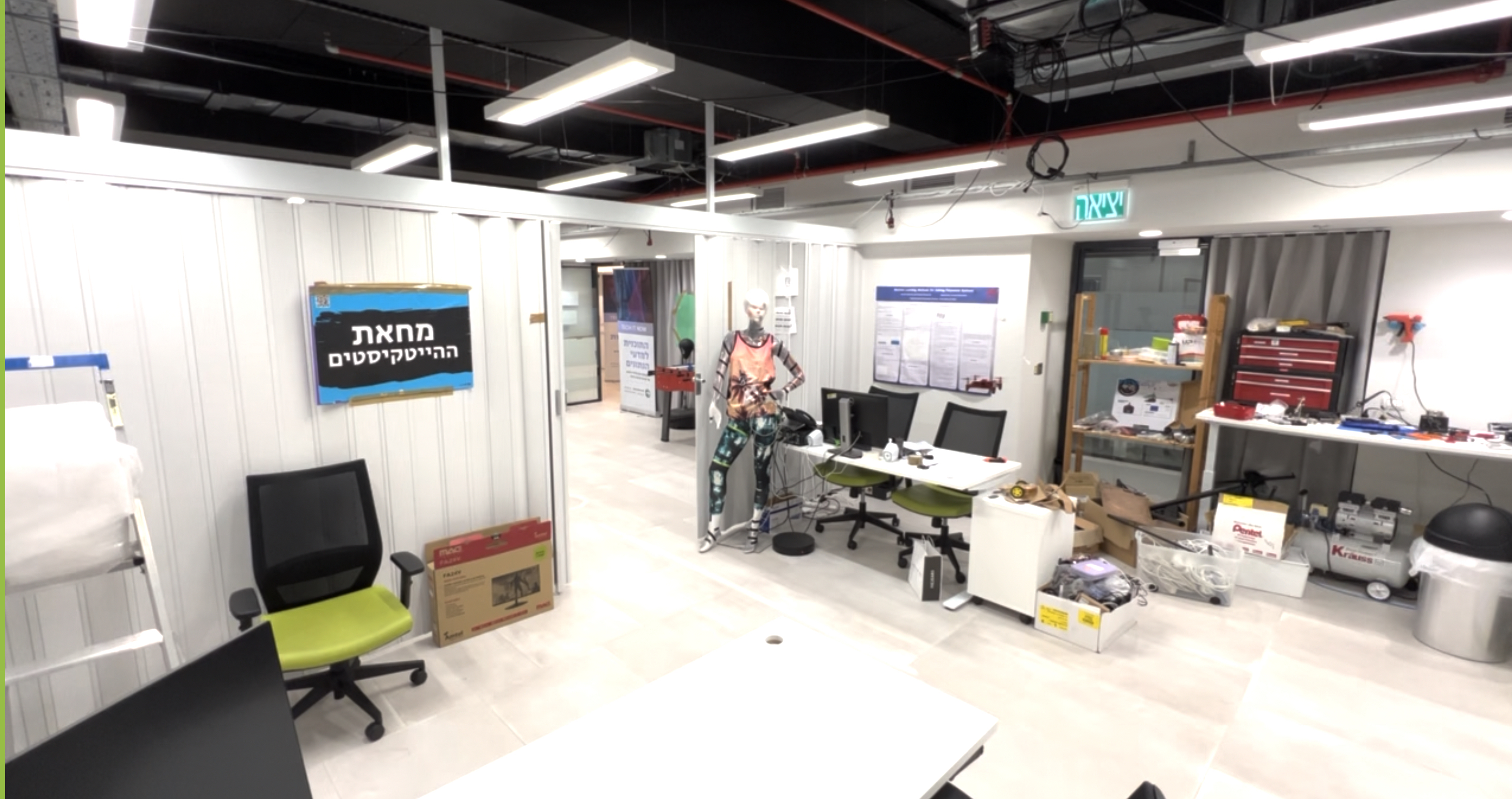}
    \caption[]%
    {{Image taken from our lab}}  
    \label{fig:fares}
\end{subfigure}
\begin{subfigure}[t]{0.33\textwidth}
 \centering
    \includegraphics[width=\textwidth,height=3cm]{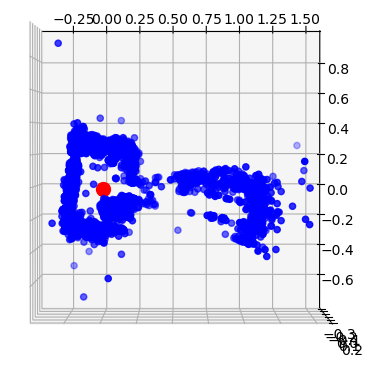}
    \caption[]%
    {{Point cloud obtained by \emph{ORB-SLAM3}}}  
    \label{fig:noisy}
\end{subfigure}
 \begin{subfigure}[t]{0.33\textwidth}  
 \centering 
        \includegraphics[width=\textwidth, height=3cm]{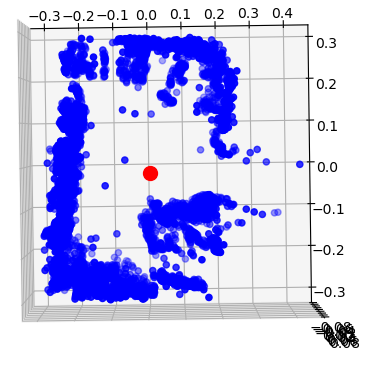}
        \caption[]%
        {Outlier removal using Algorithm~\ref{alg:square_root_approximation}}    
        \label{fig:clean}
        \end{subfigure}


        \hfill
\begin{subfigure}[t]{0.49\textwidth}
 \centering
    \includegraphics[width=0.66\textwidth,height=3cm]{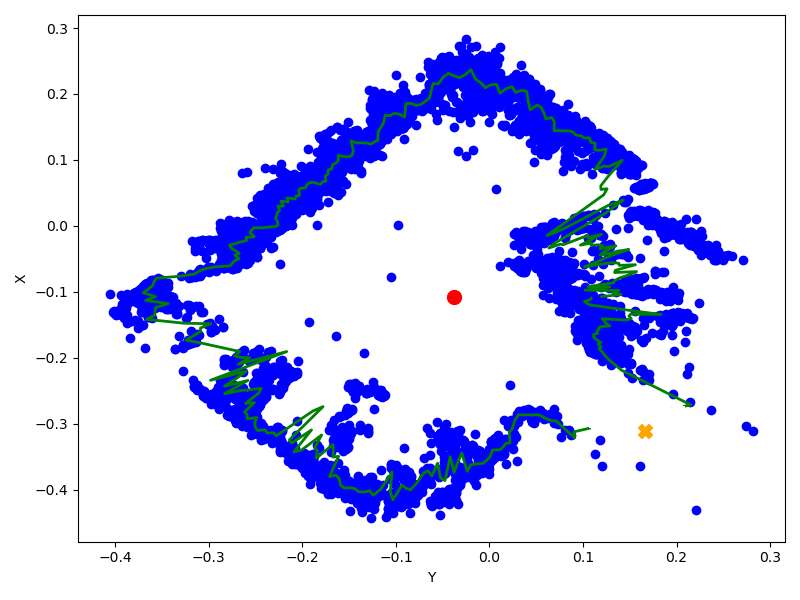}
    \caption[]%
    {{Finding an exit in a $2$D projection of the \say{clean} map}}
    \label{fig:2D}
\end{subfigure}
 \begin{subfigure}[t]{0.49\textwidth}  
 \centering 
        \includegraphics[width=0.66\textwidth, height=3cm]{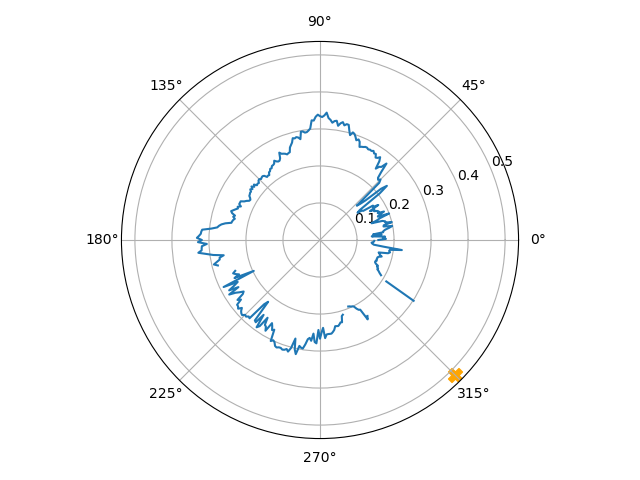}
        \caption[]%
        {{The angle of the exit provided by Algorithm~\ref{alg:exit}}}    
        \label{fig:angles}
        \end{subfigure}
        \caption{Illustration of our system on a sub-environment of our lab (Figure~\ref{fig:fares}). Blue dots are \emph{ORB-SLAM3} features, a red dot is the agent's current position, green lines are the average distances at each angle from $[0, 359]$, and the yellow $\mathbf{X}$ is the generated exit point.} 
        \label{fig:illustration}

\end{figure*}

In this section, we describe our experiments, validating and assessing the quality of our proposed exploration system.

\subsection{Experimental Settings}

\textbf{Hardware.} Our experiments were conducted using a standard HP ZBook laptop with an Intel Core i7-10750H CPU @2.60GHzx12 and 32GB of RAM. Our agent is a DJI's Tello Drone ($<90gr$, $<100\$$), equipped with a small \emph{RGB} monocular camera.

\textbf{Software.} Our system is implemented in C++ using \emph{ORB-SLAM3} algorithm for localization and mapping.

\textbf{Parameters and preprocessing.} First, in the context of the outlier removal, i.e., Algorithm~\ref{alg:square_root_approximation}, we have found that $\beta := 1$, $k := 4$ and $\lambda := 0.6$ worked best throughout our experiments. Secondly, in the context of finding a point of exit, the $2$D plane was set to be a plane that is parallel to the ground -- such a plane is generated by moving the drone in a triangular path obtaining three points while maintaining the same height from the ground. Finally, for the case of path planning, we have chosen $K := 1000$ (the number of centers) for our use of $K$-means clustering algorithm. 

\subsection{Offline Evaluation of Our Methods}
In what follows, we assess the performance of our outlier removal and our exit finding algorithm.


\textbf{Outlier Removal.} Figure~\ref{fig:noisy} illustrates the point cloud of a room generated by Tello Drone using \emph{ORB-SLAM3}. Figure~\ref{fig:clean} demonstrates the outcome of our outlier removal process applied to Figure~\ref{fig:noisy} -- our method noticeably improves map quality by effectively reducing outliers.

\textbf{Finding an Exit.}
After cleaning the \emph{ORB-SLAM3} point cloud, Algorithm~\ref{alg:exit} is applied to find an exit point. In Figure~\ref{fig:2D}, a $2$D projection of~\ref{fig:clean} is presented where the exit point is depicted by a yellow $\mathbf{X}$. Recall that this is done by processing the map into an angular-based map as described in Algorithm~\ref{alg:exit}; see Figure~\ref{fig:angles}.

\subsection{Real-Time Experiment}
Next, we describe and illustrate our real-time experiment. First, we placed a drone inside our lab (which consists of two rooms), with no prior knowledge of such an environment. The drone first lifts up, proceeding to scan the environment using~\emph{ORB-SLAM3} for feature extraction to represent the room as well as to localize the drone's relative position. 
This is done, by rotating the drone $360$ degrees compared to its initial pose, while going up and down during the rotation to maximize the number of obtained features. The drone then moves in a triangular path returning to its current position to generate a $2$D plane which will be used by Algorithm~\ref{alg:exit}.


We then employ Algorithm~\ref{alg:square_root_approximation} to remove outliers followed by using Algorithm~\ref{alg:exit} to guide the drone in exploring unknown areas. For effective path planning, we cluster the \say{clean} map using the $K$-means algorithm, and construct convex hulls for each of the clusters. Using this information, along with the exit point, as input for the RRT path planner, we obtain a path that navigates the drone from its current position to the next unknown environment. To ensure a smooth and simple path, the generated path undergoes refinement as described in subsection~\ref{sec:refinement}.

\textbf{Avoiding revisiting explored areas.} We exploit the functionality of \emph{ORB-SLAM3} to aid us in understanding whether a proposed exit point lies in an explored area. In such a case, prior to invoking Algorithm~\ref{alg:exit}, the map undergoes further processing closing off certain areas that would prevent having an exit point in explored areas. 

\begin{figure}[!htb]
    \centering
    \includegraphics[width=\linewidth]{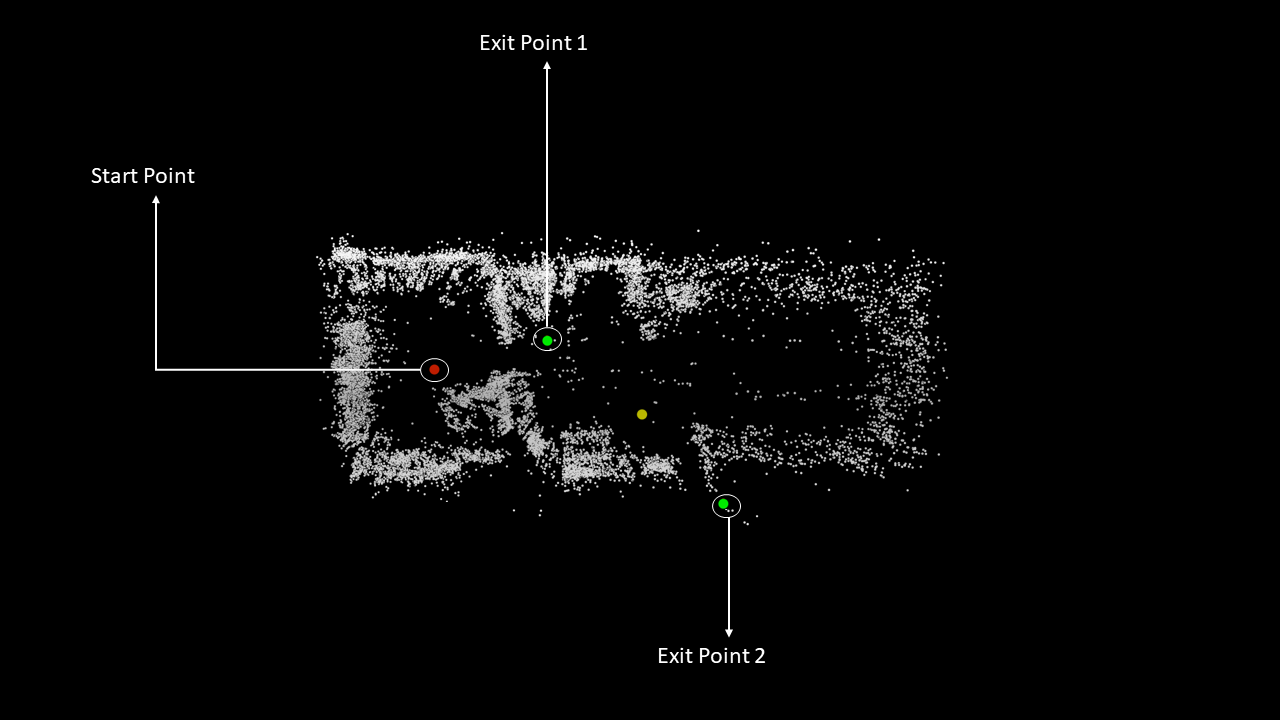}
    \caption{A composition of paths guaranteed by our proposed system that moves the agent through a pair of rooms to an area outside the lab. The yellow point is a path point, red point is the drone's initial position, and green points are exit points.}
    \label{fig:real-time-experiment}
\end{figure}

This whole procedure is iteratively applied until a previously defined criterion is satisfied, e.g., no exit points can be further explored. A result obtained during one of our real-time experiments (Figure~\ref{fig:real-time-experiment}), depicts a composition of the paths generated by our system that moves the Tello drone from one room inside the lab to outside our lab (see \say{Exit Point 2} at Figure~\ref{fig:real-time-experiment}). 
We refer the reader to our video of a similar experiment found in the supplementary material.

    
    

\section{CONCLUSIONS and FUTURE WORK}
To the best of our knowledge, our low-cost $(<100\$)$ lightweight ($<100$gr) vision featured-based monocular RGB autonomous drone navigation system is the first of its kind. We note that~\cite{khan2023vision} proposed the use of \emph{ORB-SLAM3} for navigating a Tello drone. However, no autonomous exploration system was provided.

Our system has shown efficacy throughout our experiments due to the provided outlier removal procedure, generation of points of interest (exit points), and proper transformation of \say{clean} map to ensure proper path planning using RRT.

Future work includes adapting our system to operate only on Raspberry PI~\cite{jubran2022newton}. In addition, deep learning techniques have infiltrated this field, leading to promising results, e.g.,~\cite{maalouf2023deep}. This raises an additional future work of our proposed system to involve deep learning models combined with compression methods, e.g.,~\cite{tukan2022pruning} that aim to ensure autonomous navigation under the constraint of very limited hardware such as Raspberry PI.

\section{ACKNOWLEDGMENTS}
This work was partially supported by the Center for Cyber Law \& Policy at the University of Haifa in conjunction with the Israel National Cyber Directorate in the Prime Minister’s Office. 

\addtolength{\textheight}{-12cm}   









\bibliographystyle{alpha}
\bibliography{references}

\end{document}